\documentclass[10pt, journal, twoside]{IEEEtran}

\usepackage{amsmath}
\usepackage{graphicx}
\usepackage{amsthm}
\usepackage{amssymb}
\usepackage{algorithm}
\usepackage{algpseudocode}
\usepackage{microtype}

\usepackage[caption=false]{subfig}

\newcommand{\norm}[1]{\Vert #1 \Vert}
\newcommand{\mbf}[1]{\mathbf{#1}}
\newcommand{\mc}[1]{\mathcal{#1}}

\newcommand{\ti}[1]{{\tilde{#1}}}

\DeclareMathOperator*{\argmin}{argmin}

\usepackage{adjustbox}
\usepackage{array}

\newcolumntype{R}[2]{%
    >{\adjustbox{angle=#1,lap=\width-(#2)}\bgroup}%
    l%
    <{\egroup}%
}

\begin{document}
\title{A Survey of Distributed Optimization Methods for Multi-Robot Systems}

\author{~\IEEEmembership{}Trevor~Halsted$^1$, Ola~Shorinwa$^{1}$,~\IEEEmembership{}Javier~Yu$^{2}$,~\IEEEmembership{}Mac~Schwager$^{2}$~\IEEEmembership{}%
\thanks{*This project was funded in part by DARPA YFA award D18AP00064, and NSF NRI awards 1830402 and 1925030.  The first author was supported on an NDSEG Fellowship, and the third author was supported on an NSF Graduate Research Fellowship.}
\thanks{$^{**}$ The first three authors contributed equally.}%
\thanks{$^{1}$Department of Mechanical Engineering, Stanford University, Stanford, CA 94305, USA, {\tt\small \{halsted, shorinwa\}@stanford.edu}}%
\thanks{$^{2}$Department of Aeronautics and Astronautics, Stanford University, Stanford, CA 94305, USA
        {\tt\small \{javieryu, schwager\}@stanford.edu}}%
        }
    
\maketitle 
	
\begin{abstract}
Distributed optimization consists of multiple computation nodes working together to minimize a common objective function through local computation iterations and network-constrained communication steps.  In the context of robotics, distributed optimization algorithms can enable multi-robot systems to accomplish tasks in the absence of centralized coordination. We present a general framework for applying distributed optimization as a module in a robotics pipeline.  We survey several classes of distributed optimization algorithms and assess their practical suitability for multi-robot applications.  We further compare the performance of different classes of algorithms in simulations for three prototypical multi-robot problem scenarios.  The Consensus Alternating Direction Method of Multipliers (C-ADMM) emerges as a particularly attractive and versatile distributed optimization method for multi-robot systems.
\end{abstract}

\begin{IEEEkeywords}
	distributed optimization, multi-robot systems
\end{IEEEkeywords}

\section{Introduction}   
Distributed optimization is the problem of minimizing a joint objective function consisting of a sum of several local objective functions, each corresponding to a computational node. While distributed optimization has been a longstanding topic of research in the optimization community (e.g., \cite{rockafellar1976monotone, tsitsiklis1984problems}), its usage in robotics is limited to only a handful of examples. However, distributed optimization techniques have important implications for multi-robot systems, as we can cast many of the key tasks in this area, including cooperative estimation \cite{ola2020targettracking}, multi-agent learning \cite{wai2018multi}, and collaborative motion planning \cite{bento2013message}, as distributed optimization problems. The distributed optimization formulation offers a flexible and powerful paradigm for deriving efficient and distributed algorithms for many multi-robot problems. In this survey, we provide an overview of the distributed optimization literature, contextualize it for applications in robotics, discuss best practices for applying distributed optimization to robotic systems, and evaluate several algorithms in simulations for fundamental robotics problems.  

We specifically consider optimization problems over real-valued decision variables (we do not consider discrete optimization, i.e., integer programs or mixed integer programs), and we assume that the robots communicate over a mesh network without central coordination.  Often times, in the context of robotics, optimization is performed on-line within another iterative algorithm, e.g. for planning and control---as in Model Predictive Control (MPC), or perception---as in on-line Simultaneous Localization and Mapping (SLAM).  Hence one can understand the algorithms in this paper as being run within a single time step of another iterative algorithm. 

In evaluating the usefulness of distributed optimization algorithms for multi-robot applications, we focus on methods that permit robots to use local robot-to-robot communication to compute solutions that are of the same quality as those that would have been obtained were all the robots' observations available on one central computer.  For convex problems, the robots all obtain a common globally optimal solution using distributed optimization.  More generally, for non-convex problems, all robots obtain a common solution which may be locally optimal, but still of the same quality as that which would have been obtained by a single computer with all problem data.

\begin{figure}[t!]
\centering
\subfloat[Optimization by one robot yields the solution given only that robot's observations.]{%
  \includegraphics[width=0.8\columnwidth]{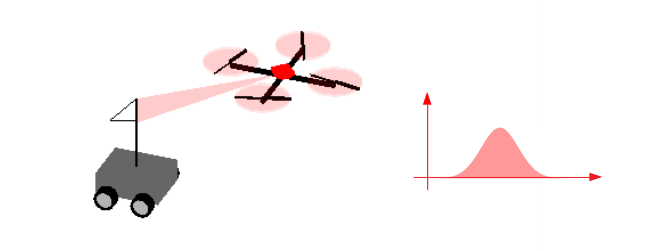}%
}

\subfloat[Using distributed optimization, each robot obtains the optimal solution resulting from all robots' observations.]{%
  \includegraphics[width=0.8\columnwidth]{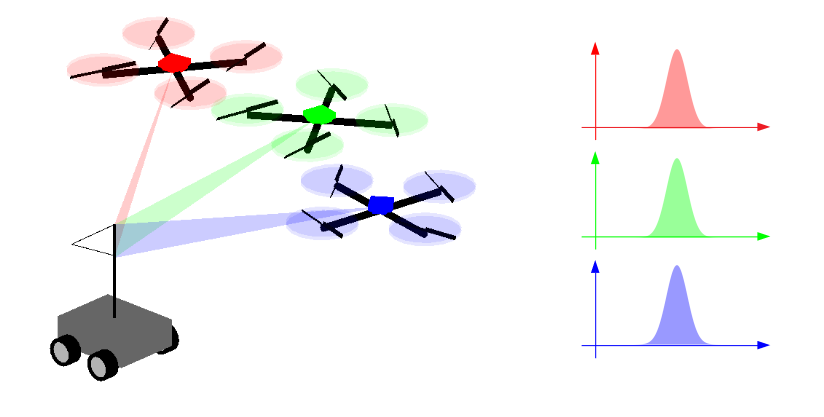}%
  
}

\caption{A motivation for distributed optimization: consider an estimation scenario in which a robot seeks to localize a target given sensor measurements. The robot can compute an optimal solution given \emph{only its observations}, as represented in (a). By using distributed optimization techniques, each robot in a networked system of robots can compute the optimal solution \emph{given all robots' observations} without actually sharing individual sensor models or measurements with one another, as represented in (b).}

\end{figure}
In this survey, we provide a taxonomy of the range of different algorithms for performing distributed optimization based on their defining mathematical characteristics, and categorize the distributed optimization algorithms into three classes: distributed gradient descent, distributed sequential convex programming, and distributed extensions to the alternating direction method of multipliers (ADMM). We do not discuss zeroth-order methods for distributed optimization \cite{hajinezhad2017zeroth, hajinezhad2019zone, hajinezhad2019perturbed}.
    
\textbf{Distributed Gradient Descent:} The most common class of distributed optimization methods is based on the idea of averaging local gradients computed by each computational node to perform an approximate gradient descent \cite{tsitsiklis1986}, and in this work we refer to them as distributed gradient descent (DGD) methods. Accelerated DGD algorithms, like \cite{shi2015extra}, offer significantly faster convergence rates over basic DGD. Furthermore, DGD methods have been shown to converge to the optimal solution on non-differentiable convex functions with subgradients \cite{nedic2009} and with a push-sum averaging scheme \cite{nedic2014distributed}, making them well suited for a broad range of applications. 
    
\textbf{Distributed Sequential Convex Programming:} Sequential Convex Optimization is a common technique in centralized optimization that involves minimizing a sequence of convex approximations to the original (usually non-convex) problem.  Under certain conditions, the sequence of sub-problems converges to a local optimum of the original problem.  Newton's method and the 
Broyden–Fletcher–Goldfarb–Shanno (BFGS) method are common examples.  The same concepts are used by a number of distributed optimization algorithms, and we refer to these algorithms as distributed sequential convex programming methods. Generally, these methods use consensus techniques to construct the convex approximations of the joint objective function. One example is the Network Newton method \cite{Mokhtari2015} which uses consensus to approximate the inverse Hessian of the objective to construct a quadratic approximation of the joint problem. The NEXT family of algorithms \cite{di2016next} provides a flexible framework which can utilize a variety of convex surrogate functions to approximate the joint problem, and is specifically designed to optimize non-convex objective functions. 
    
\textbf{Alternating Direction Method of Multipliers (ADMM):} The last class of algorithms covered in this paper are based on the alternating direction method of multipliers (ADMM) \cite{rockafellar1976monotone}.  ADMM works by minimizing the augmented Lagrangian of the optimization problem using alternating updates to the primal and dual variables \cite{boyd2011distributed}.  The method is naturally amenable to constrained problems.  The original method is distributed, but not in the sense we consider in this survey.  Specifically, the original ADMM requires a central computation hub to collect all local primal computations from the nodes to perform a centralized dual update step.  ADMM was first modified to remove this requirement for a central node in \cite{mateos2010distributed}, where it was used for distributed signal processing.  The algorithm from \cite{mateos2010distributed} has since become known as Consensus ADMM (C-ADMM), although that paper does not introduce this terminology.  C-ADMM was later shown to have linear convergence rates on all strongly convex distributed optimization problems \cite{shi2014linear}.  We find C-ADMM outperforms the other optimization algorithms that we tested in our three problem scenarios in terms of convergence speed, as well as computational and communication efficiency. In addition, we find that it is less sensitive to the choice of hyper-parameters than the other methods we tested.  Therefore, C-ADMM emerges as an attractive option for problems in multi-robot systems. 
    
Many problem features affect the convergence rates of each of these methods, and convergence guarantees are often qualified by the convexity of the underlying joint optimization problem. Essential to applying these algorithms is understanding their limitations and performance trade-offs. For instance, some methods are better suited for handling problems with constraints, but have a higher computational complexity or communication overhead.
    
Numerical results in this survey implement select algorithms from each of our taxonomic classes, and use them to solve three fundamental problems in multi-robot systems: cooperative target tracking, planning for coordinated package delivery, and cooperative multi-robot range-only mapping. Typically, distributed optimization algorithms are designed to solve convex problems, but often the optimization problems that arise in robotics applications are non-convex. The goal of these numerical simulations is to compare the performance of these algorithms not only in convex optimization where their convergence is often guaranteed, but also in non-convex problems where it is not. 

\subsection{Relevance to Robotics}    
While the field of distributed optimization is well-developed, its application to robotics problems remains nascent. As a result, many existing distributed optimization methods do not cater to the specific challenges that arise in robotics problems. Generally, robotics problems involve constrained non-convex optimization problems, an area not explored by many distributed optimization methods. Further, robots typically have limited access to significant computation and communication resources, placing greater importance on efficient optimization methods with low overhead. Many distributed optimization methods do not consider these issues with the assumption that agents have access to sufficient and reliable computation and communication infrastructure. For relevance to robotics problems, we specifically note methods that work on constrained problems and quantify the relative computation and communication costs incurred by distributed optimization methods.
    
In much of the literature, distributed optimization algorithms are compared on a convergence per iteration basis (per update to the decision variable). However, this scheme often obfuscates critical information for applications to robotics like local computation time, parameter sensitivity, and communication overhead. As part of the numerical results, this survey presents a more structured approach to analyzing the strengths and weaknesses of these algorithms in terms of metrics that matter for robotics research.
    
    \subsection{Existing Surveys}
    A number of other recent surveys on distributed optimization exist, and provide useful background when working with the algorithms covered in this survey. The survey \cite{molzahn2017survey} covers applications of distributed optimization for applications in distributed power systems, and \cite{nedic2018distributed} focuses on the application of predominately first-order methods to solving problems in multi-agent control. The article \cite{nedic2018network} broadly addresses communication-computation trade-offs in distributed optimization, again focused mainly on first-order methods. Another survey \cite{chang2020distributed}, covers exclusively non-convex optimization in both batch and data-streaming contexts, but again only analyzes first-order methods. Finally, \cite{yang2019survey} covers a wide breadth of distributed optimization algorithms with a variety of assumptions focusing exclusively on convex optimization tasks. This survey differs from all of these in that it is specifically targeting optimization applications in robotics, and provides a condensed taxonomic overview of useful methods for these applications.
    
    Other useful background material can be found for distributed computation \cite{bertsekas1989parallel} \cite{lynch1996distributed}, and on multi-robot systems in \cite{distctrlrobotnetw} \cite{mesbahi2010graph}. 
    
    \subsection{Contributions}
    This survey paper has four primary objectives:
    \begin{enumerate}
        \item Provide a unifying formulation for the general distributed optimization problem.
        \item Develop a taxonomy for the different classes of distributed optimization algorithms.
        \item Compare the performance of distributed optimization algorithms for applications in robotics with varying levels of difficulty.
        \item Propose open research problems in distributed optimization for robotics.
    \end{enumerate}
    
    \subsection{Organization}
    In Section \ref{sec:problem_formulation}, we present the general formulation for the distributed optimization problem, and give insight into the basic assumptions typically made while developing distributed optimization algorithms. Sections \ref{sec:DistGradDesc} - \ref{sec:ADMM} provide greater detail on our algorithm classifications, and include details for representative algorithms. Finally, Section \ref{sec:simulations} gives performance comparisons for select algorithms on a range of different optimization problems specifically highlighting communication versus computation trade-offs.  In Sec.~\ref{sec:open_problems} we discuss open research problems in distributed optimization applied to multi-robot systems and robotics in general, and we offer concluding remarks in Sec.~\ref{sec:conclusion}.
    
	\section{Problem Formulation}
	\label{sec:problem_formulation}
	
	In distributed optimization in multi-robot systems, robots perform communication and computation steps to minimize some global cost function. We focus on problems in which the robots' exchange of information must respect the topology of an underlying distributed communication graph.  This communication graph, denoted as $\mc{G} = (\mc{V}, \mc{E})$, consists of vertices $\mc{V} = \{1, \dots, n\}$ and edges $\mc{E} \subseteq \mc{V} \times \mc{V}$ over which pairwise communication can occur. In general, we assume that the communication graph is connected (there exists some path of edges from any robot $i$ to any other robot $j$) and undirected (if $i$ can send information to $j$, then $j$ can send information to $i$) but place no other assumptions on its structure.
	
    In the general distributed optimization formulation, each robot $i \in \mc{V}$ can compute its local cost function $f_i(x)$ but cannot directly compute the local cost functions of the other robots. The robots' collective objective is to minimize the global cost function
	\begin{equation}
	f(x) = \sum_{i \in \mc{V}} f_i(x) \label{eq:global_cost_function}
	\end{equation}
	despite the limitations on the local information at each robot.  The problem in \eqref{eq:global_cost_function} is separable in that it consists of the sum of the local cost functions.
	
	In addition to local knowledge of the local objective functions, we assume that constraints on the optimization variable are only known locally as well.  Therefore, the general form of the global cost function is	
	\begin{align}
	\label{eq:general_problem}
	\begin{split}
	\min_x \:&\sum_{i \in \mc{V}} f_i(x)\\
	\text{subject to }&g_i(x) = 0 \quad \forall i \in \mc{V}
	\\& h_i(x) \le 0 \quad \forall i \in \mc{V}.
	\end{split}
	\end{align}
	In this paper, we evaluate distributed algorithms on several classes of the separable optimization problem, including unconstrained linear least-squares, constrained convex, and constrained non-convex optimization problems.
	
	Before describing the specific algorithms that solve distributed optimization problems, we first consider the general framework that all of these approaches share.  Each algorithm progresses over discrete iterations $k = 0, 1, \dots$ until convergence. Besides assuming that each robot has the sole capability of evaluating its local cost function $f_i$, we also distinguish between the ``private'' variables $\mc{P}_i^{(k)}$ that the robot computes at each iteration $k$ and the ``public'' variables $\mc{Q}_i^{(k)}$ that the robot communicates to its neighbors.  Each algorithm also involves parameters $\mc{R}_i^{(k)}$, which generally require coordination among all of the robots but can typically be assigned before deployment of the system.  Algorithm \ref{alg:GenDistOpt} describes the general framework for distributed optimization, in which each iteration $k$ involves a communication step and a computation step.
	
\begin{algorithm}[t]
	\caption{General Distributed Optimization Framework}\label{alg:GenDistOpt}
	\begin{algorithmic}[1]
	\Function{DistOpt}{$f_i, \mc{P}_i^{(0)}, \mc{Q}_i^{(0)}, \mc{R}_i^{(0)} \: \forall i \in \mc{V}$}
	\State $k \gets 0$
	\While{stopping criterion is not satisfied}
	    \For{$i \in \mc{V}$} (in parallel)
	    \State Communicate $\mc{Q}_i^{(k)}$ to all $j \in \mc{N}_i$
	    \State Receive $\mc{Q}_j^{(k)}$ from all $j \in \mc{N}_i$
	    \State Compute $\mc{P}_i^{(k+1)}, \mc{Q}_i^{(k+1)}, \mc{R}_i^{(k+1)}$
	    \EndFor
	    \State $k \gets k + 1$
	\EndWhile
	\EndFunction
	\end{algorithmic}
    \end{algorithm}

    In order for each robot to find the joint solution to \eqref{eq:general_problem} in a distributed manner, the update steps in algorithm \ref{alg:GenDistOpt} must allow the robots to cooperatively balance the costs accrued in each local cost function. From the perspective of a single robot, the update equations represent a trade-off between optimality of its individual solution based on its local cost function and agreement with its neighbors. In this paper, we distinguish between two distinct perspectives on how this agreement is achieved. 
    
	In the following sections, we discuss three broad classes of distributed optimization methods, including distributed gradient descent, distributed sequential convex programming, and the alternating direction method of multipliers. In distributed gradient descent methods, the robots update their local variables using their local gradients and obtain the same solution of the global problem through consensus on their local variables. Distributed sequential convex programming methods take a similar approach but involve the problem Hessian, which provides information on the function curvature when updating the local variables. 
	
	The alternating direction method of multipliers takes a different approach by explicitly enforcing agreement between the robots' local variables before relaxing these constraints using the problem's Lagrangian. In this approach, we consider the reformulation of \eqref{eq:global_cost_function} to distinguish between each robots estimate of the decision variable $x_i$:
    \begin{align}
	\label{eq:general_problem_with_agreement}
	\begin{split}
	\min_{x_i \forall i \in \mc{V}} \:&\sum_{i \in \mc{V}} f_i(x_i)\\
	\text{subject to }& x_i = x_j \quad \forall (i, j) \in \mc{E} \\&g_i(x_i) = 0 \quad \forall i \in \mc{V}
	\\& h_i(x_i) \le 0 \quad \forall i \in \mc{V}
	\end{split}
	\end{align}
	where the agreement constraints are only enforced between neighboring robots. We discuss distributed gradient descent before proceeding with a discussion on distributed sequential convex programming and the alternating direction method of multipliers.

	\section{Distributed Gradient Descent}\label{sec:DistGradDesc}
	The optimization problem in \eqref{eq:general_problem} (in its unconstrained form) can be solved through gradient descent where the optimization variable is updated using
	\begin{equation}
	    x^{(k+1)} = x^{(k)} - \alpha^{(k)} \nabla f(x^{(k)}) \label{eqn:cent_grad_desc}
	\end{equation}
	with $\nabla f(x^{(k)})$ denoting the gradient of the objective function, ${\nabla f(x) = \sum _{i \in \mathcal{V}} \nabla f_i(x)}$, given some scheduled step-size $\alpha^{(k)}$. Inherently, computation of $\nabla f(x^{(k)})$ requires knowledge of the local objective functions or gradients by all robots in the network which is infeasible in many problems.
	
	\textit{Distributed Gradient Descent} (DGD) methods extend the centralized gradient scheme to the distributed setting where robots communicate locally without necessarily having knowledge of the local objective functions or gradients of all robots. In DGD methods, each robot updates its local variable using a weighted combination of the local variables or gradients of its neighbors according to the weights specified by a weighting matrix $W$, allowing for the dispersion of information on the objective function or its gradient through the network. In general, the weighting matrix $W$ reflects the topology of the communication network, with non-zero weights existing between pairs of neighboring robots. The weighting matrix exerts a significant influence on the convergence rates of DGD methods, and thus, an appropriate choice of these weights are required for convergence of DGD methods.
	
    Many DGD methods use a doubly stochastic matrix $W$, a row-stochastic matrix $A$ \cite{mai2019distributed}, or a column-stochastic matrix $B$, depending on the model of the communication network considered, while other methods use a push-sum approach. In addition, many methods further require symmetry of the doubly stochastic weighting matrix with ${W^{\top}\mbf{1} = \mbf{1}}$ and ${W = W^{\top}}$.
	
	The use of doubly stochastic matrices is supported by a rich body of work on the convergence of consensus algorithms of this form. (As observed in several works including \cite{lobel2010distributed}, the standard average consensus problem is equivalent to using \eqref{eqn:basic_DGD} to solve the distributed optimization problem \eqref{eq:general_problem} when the local cost functions $f_i$ are constant). 
	
	DGD methods generally required decreasing step sizes for convergence to an optimal solution of the problem which negatively impacts their convergence rates; however, these methods have been modified to develop gradient descent methods with fixed step-sizes. Next, we discuss these distributed gradient descent variants.

	\subsection{Decreasing Step-Size DGD}
	
	\begin{algorithm}[t]
	\caption{Distributed Gradient Descent (DGD)}\label{alg:DistGradDesc}
    \textbf{Private variables:} $\mc{P}_i = \emptyset$
    \\\textbf{Public variables:} $\mc{Q}_i^{(k)} = x_i^{(k)}$
    \\\textbf{Parameters:} $\mc{R}_i^{(k)} = (\alpha^{(k)}, w_i)$
    \\\textbf{Update equations:} \begin{align*}x_i^{(k+1)} &= w_{ii} x_i^{(k)} + \sum_{j \in \mathcal{N}_i} w_{ij} x_j^{(k)} - \alpha^{(k)} \nabla f_i(x_i^{(k)})\\
    \alpha^{(k+1)} &= \frac{\alpha^{(0)}}{\sqrt{k}}\end{align*}
    \end{algorithm}
    
	Tsitsiklis introduced a model for DGD in the 1980s in \cite{tsitsiklis1984problems} and \cite{tsitsiklis1986} (see also \cite{bertsekas1989parallel}). The works of Nedi\'{c} and Ozdaglar in \cite{nedic2009} revisit the problem, marking the beginning of interest in consensus-based frameworks for distributed optimization over the recent decade.  This basic model of DGD consists of an update term that involves consensus on the optimization variable as well as a step in the direction of the local gradient for each node:
	\begin{equation}
	    x_i(k+1) = w_{ii} x_i(k) + \sum_{j \in \mathcal{N}_i} w_{ij} x_j(k) - \alpha_i(k) \nabla f_i(x_i(k)) \label{eqn:basic_DGD}
	\end{equation}
	where robot $i$ updates its variable using a weighted combination of its neighbors' variables determined by the weights $w_{ij}$ with $\alpha_{i}(k)$ denoting its local step-size at iteration $k$. 
	
	For convergence to the optimal joint solution, these methods require the step-size to asymptotically decay to zero. As proven in \cite{lobel2010distributed}, scheduling the step-size according to the rules $\sum_{k = 1}^\infty \alpha_i(k) = \infty$ and $\sum_{k = 1}^\infty \alpha_i(k)^2 < \infty$ with all robots' step-sizes equal guarantees the asymptotic convergence of the robots' optimization variables to the optimal joint solution, given the standard assumptions of a connected network, properly chosen weights, and bounded (sub)gradients.  Alternatively, the choice of a constant step-size for all timesteps only guarantees convergence of each robot's iterates to a neighborhood of the optimal joint solution.
	
    Algorithm \ref{alg:DistGradDesc} summarizes the update step for the decreasing step-size gradient descent method in Algorithm \ref{alg:GenDistOpt} with the step-size $\alpha^{(k+1)} = \frac{\alpha^{(0)}}{\sqrt{k}}$.
    
    
	An alternative approach to the consensus term in \eqref{eqn:basic_DGD} uses \textit{push-sum} consensus introduced in the context of gossip-based distributed algorithms \cite{kempe2003gossip}. This approach involves two parallel consensus steps to circumvent the need for a doubly stochastic matrix which is replaced by a row-stochastic matrix $A$. The push-sum technique is also explored in \cite{olshevsky2009convergence, olshevsky2018robust, nedic2014distributed, benezit2010weighted} in distributed gradient methods. In general, using a row-stochastic weighting matrix $A$ in place of a doubly stochastic $W$ would result in the consensus step becoming weighted according to the relative degrees of each robot in the communication graph.  The push-sum approach introduces a second variable by which the robots keep track of their relative degrees and ``unweight'' the consensus term. The extra communication cost of this second variable comes with the benefit of extending consensus behavior to networks with asynchronous or directed communication.
	
	Noting the sub-linear convergence rates of decreasing step-size DGD, some approaches have applied accelerated gradient schemes for improved convergence speed \cite{jakovetic2014fast}.
	
    \subsection{Fixed Step-Size DGD}
    
 \begin{algorithm}[t]
	\caption{EXTRA}\label{alg:EXTRA}
    \textbf{Private variables:} $\mc{P}_i^{(k)} = \emptyset$
    \\\textbf{Public variables:} $\mc{Q}_i^{(k)} = \left(x_i^{(k)}, x_i^{(k-1)}\right)$
    \\\textbf{Parameters:} $\mc{R}_i^{(k)} = \left(\alpha, w_i\right)$
    \\\textbf{Update equations:} \begin{align*}
    x_i^{(k+1)} &= \sum_{j \in \mc{N}_i \cup \{i\}}w_{ij} \left(x_j^{(k)} - \frac{1}{2} x_j^{(k-1)} \right)\cdots\\&\qquad - \alpha \left[\nabla f_i\left(x_i^{(k)}\right) - \nabla f_i\left(x_i^{(k-1)}\right)\right]
    \end{align*}
    \end{algorithm}

	Although decreasing step-size DGD methods converge to an optimal joint solution, the requirement of a decaying step-size reduces the convergence speed of these methods. Fixed step-size methods address this limitation by eliminating the need for decreasing step-sizes while retaining convergence to the optimal joint solution. The EXTRA algorithm introduced by Shi \textit{et al.} in \cite{shi2015extra} uses a fixed step-size while still achieving exact convergence. EXTRA replaces the gradient term with the difference in the gradients of the previous two iterates. Because the contribution of this gradient difference term decays as the iterates converge to the optimal joint solution, EXTRA does not require the step-size to decay in order to settle at the exact global solution. Algorithm \ref{alg:EXTRA} describes the update steps for the local variables of each robot in EXTRA. EXTRA achieves linear convergence \cite{yuan2016convergence}, and a variety of DGD algorithms have since offered improvements on its linear rate \cite{daneshmand2018second}. Many other works with fixed step-sizes involve variations on the variables updated using consensus and the order of the update steps, including DIGing \cite{nedic2017achieving, nedic2017digging}, NIDS \cite{li2019decentralized}, Exact Diffusion \cite{yuan2018exact, yuan2018exact2}, and \cite{qu2017harnessing, xin2019distributed}. These approaches which generally require the use of doubly stochastic weighting matrices have been extended to problems with row-stochastic or column-stochastic matrices \cite{saadatniaki2018optimization, xi2017add, xin2018linear, xi2018linear} and push-sum consensus \cite{zeng2017extrapush} for distributed optimization in directed networks. Several works offer a synthesis of various fixed step-size DGD methods, noting the similarities between the fixed step-size DGD methods. Under the canonical form proposed in \cite{sundararajan2019canonical}, these algorithms and others differ only in the choice of several constant parameters. We provide this unifying template for fixed step-size DGD methods in Algorithm \ref{alg:FixedStepSizeDGD}, included in the Appendix \ref{sec:FixedStepSize}. Jakoveti\'{c} also provides a unified form for various fixed-step-size DGD algorithms in \cite{jakovetic2018unification}. Some other works consider accelerated variants using Nesterov gradient descent \cite{qu2019accelerated, xin2019distributedNesterov, qu2019accelerated, lu2020nesterov}. 
	
	In general, DGD methods address unconstrained distributed convex optimization problems, but these methods have been extended to non-convex problems \cite{tatarenko2017non} and constrained problems using projected gradient descent \cite{ram2010distributed, bianchi2012convergence, johansson2010randomized}.
	
	Some other methods perform dual-ascent on the dual problem of \eqref{eq:general_problem} \cite{maros2018panda, maros2020geometrically, maros2019eco, seaman2017optimal} where the robots compute their local primal variables from the related minimization problem using their dual variables. These methods require doubly stochastic weighting matrices but allow for time-varying communication networks. In \cite{lan2020communication}, the robots perform a subsequent proximal projection step to obtain solutions which satisfy the problem constraints.
	
	\subsection{Distributed Dual Averaging}
	
    \begin{algorithm}[t]
	\caption{Distributed Dual Averaging (DDA)}\label{alg:DistDualAvg}
    \textbf{Private variables:} $\mc{P}_i = z_i^{(k)}$
    \\\textbf{Public variables:} $\mc{Q}_i^{(k)} = x_i^{(k)}$
    \\\textbf{Parameters:} $\mc{R}_i^{(k)} = \left(\alpha^{(k)}, w_i, \phi(\cdot)\right)$
    \\\textbf{Update equations:}
    \begin{align*}
    z_i^{(k+1)} &= \sum_{j \in \mc{N}_i} w_{ij} z_j^{(k)} + \nabla f_i\left(x_i^{(k)}\right)\\
    x_i^{(k+1)} &= \argmin_{x \in \mc{X}} \left\{x^\top z_i^{(k+1)} + \frac{1}{\alpha^{(k)}} \phi(x) \right\}
    \end{align*}
    \end{algorithm}
    
    
    
    Dual averaging first posed in \cite{nesterov2009primal}, and extended in \cite{xiao2010dual}, takes a similar approach to gradient descent methods in solving the optimization problem in \eqref{eq:general_problem}, with the added benefit of providing a mechanism for handling problem constraints through a projection step in like manner as projected gradient descent methods. However, the original formulations of the dual averaging method requires knowledge of all components of the objective function or its gradient which is unavailable to all robots. The Distributed Dual Averaging method (DDA) circumvents this limitation by modifying the update equations using a doubly stochastic weighting matrix to allow for updates of each robot's variable using its local gradients and a weighted combination of the variables of its neighbors \cite{duchi2011dual}.
    
    Similar to decreasing step-size DGD methods, distributed dual averaging requires a sequence of decreasing step-sizes to converge to the optimal solution. Algorithm \ref{alg:DistDualAvg} provides the update equations in the DDA method, along with the projection step which involves a proximal function $\psi(x)$, often defined as $\frac{1}{2}\norm{x}_2^2$. After the projection step, the robot's variable satisfy the problem constraints described by the constraints set $\mc{X}$.
    
    
    Some of the same extensions made to DGD have been studied for DDA including analysis of the algorithm under communication time delays \cite{tsianos2011distributed}, and replacement of the doubly stochastic weighting matrix with push-sum consensus \cite{tsianos2012push}.

	\subsection*{Applications of DGD}
    Distributed gradient descent methods have found notable applications in robot localization from relative measurements problems \cite{dang2016decentralized, alwan2015distributed} including in networks with asynchronous communication \cite{todescato2015distributed}. DGD has also been applied to optimization problems on manifolds including $SE(3)$ localization \cite{tron2009distributed, tron2011distributed, tron2012distributed, tron2014distributed}, synchronization problems \cite{sarlette2009consensus}, and formation control in $SO(3)$ \cite{oh2013formation, oh2018distributed}. Other works \cite{montijano2014efficient} employ DGD along with a distributed simplex method \cite{burger2012distributed} to obtain an optimal assignment of the robots to a desired target formation. In pose graph optimization, DGD has been employed through majorization minimization schemes which minimize an upper-bound of the objective function \cite{fan2020majorization} and using gradient descent on Riemannian manifolds \cite{tian2020asynchronous, knuth2013collaborative}, and \cite{tian2019block} (block-coordinate descent).
    
    Online problems are a field of particular interest for distributed optimization algorithms, and a number of works adapted DDA for online scenarios,  \cite{hosseini2013online, shahrampour2013exponentially}, with several implemented in scenarios with time varying communication topology \cite{hosseini2016online}, \cite{lee2017stochastic}. The push-sum variant of dual averaging has also been used for distributed training of deep-learning algorithms, and has been shown to be useful in avoiding pitfalls of other distributed training frameworks including communication deadlocks and asynchronous update steps \cite{tsianos2012application}.

	\section{Distributed Sequential Convex Programming}\label{sec:DistSeqCvx}
	\subsection{Approximate Newton Methods}
    Newton's method, and its variants, are commonly used for solving convex optimization problems, and provide significant improvements in convergence rate when second-order function information is available \cite{boyd2004convex}. While the distributed gradient descent methods exploit only information on the gradients of the objective function, Newton's method uses the Hessian of the objective function, providing additional information on the function's curvature which can improve convergence. To apply Newton's method to the distributed optimization problem in \eqref{eq:general_problem}, the Network Newton-$K$ (NN-$K$) algorithm \cite{Mokhtari2015} takes a penalty-based approach which introduces consensus between the robots' variables as components of the objective function. 
    The NN-$K$ method reformulates the constrained form of the distributed problem in \eqref{eq:general_problem} as the following unconstrained optimization problem:  
    \begin{equation}
        \label{eq:penalty_distributed_problem}
	    \min_{x \forall i \in \mathcal{V}} \: \alpha \sum_{i \in \mc{V}} f_i(x_{i}) + 
	    x_i^{\top} ( \sum_{j \in \mc{N} \cup \{i\}} \bar{w}_{ij}x_{j})
    \end{equation}
    where $\bar{W} = I - W$, and $\alpha$ is a weighting hyper parameter.
    
     \begin{algorithm}[t]
	\caption{Network Newton-$K$ (NN-$K$)}\label{alg:NetNewtonPa}
    \textbf{Private variables:} $\mc{P}_i^{(k)} = \left(g_i^{(k)}, D_i^{(k)}\right)$
    \\\textbf{Public variables:} $\mc{Q}_i = \left(x_i^{(k)}, d_i^{(k+1)}\right)$
    \\\textbf{Parameters:} $\mc{R}_i = \left(\alpha, \epsilon, K, \Bar{w}_i \right)$
    \\\textbf{Outer update equations:}
    \begin{align*}
    D_i^{(k+1)} &= \alpha \nabla^2 f_i(x_i^{(k)}) + 2\Bar{w}_{ii} I\\
    g_i^{(k+1)} &= \alpha \nabla f_i(^{(k)}) + \sum_{j \in \mc{N}_i \cup \{i\}} \Bar{w}_{ij} x_j^{(k)} \\
    d_i^{(0)} &= -\left(D_i^{(k+1)}\right)^{-1} g_i\\ &\quad\rhd\textit{compute $d_i^{(k+1)}$ via $K$ inner updates} \\
    x_i^{(k + 1)}  &= x_i^{(k)} + \epsilon \: d_i^{(k+1)}
    \end{align*}
    \textbf{Inner update equations:} (Hessian approximation)
    \begin{align*}
        d_i^{(p + 1)} &= \left(D_i^{(k)}\right)^{-1} \left[\bar{w}_{ii} d_i^{(p)} - g_i^{(k+1)} - \sum_{j \in \mc{N}_i} \Bar{w}_{ij} d_j^{(p)} \right]
    \end{align*}
    \end{algorithm}
    
    
    
    However, the Newton descent step requires computing the inverse of the joint problem's Hessian which cannot be directly computed in a distributed manner as its inverse is dense. To allow for distributed computation of the Hessian inverse, NN-$K$ uses the first $K$ terms of the Taylor series expansion ${(I - X)^{-1} = \sum_{j=0}^{\infty}X^{j}}$ to compute the approximate Hessian inverse, as introduced in \cite{zargham2013accelerated}. Approximation of the Hessian inverse comes at an additional communication cost, and requires an additional $K$ communication rounds per update of the primal variable. Algorithms \ref{alg:NetNewtonPa} summarizes the update procedures in the NN-$K$ method in which $\epsilon$ denotes the step-size for the Newton's step. As presented in Algorithm \ref{alg:NetNewtonPa}, NN-$K$ proceeds through two sets of update equations: an outer set of updates that initializes the Hessian approximation and computes the decision variable update and an inner Hessian approximation update; a communication round precedes the execution of either set of update equations.  Increasing $K$, the number of intermediary communication rounds, improves the accuracy of the approximated Hessian inverse at the cost of increasing the communication cost per primal variable update. 
    
    A follow-up work optimizes a quadratic approximation of the augmented Lagrangian of the general distributed optimization problem \eqref{eq:general_problem} where the primal variable update involves computing a $P$-approximate Hessian inverse to perform a Newton descent step, and the dual variable update uses gradient ascent \cite{Mokhtari2016}. The resulting algorithm Exact Second Order Method (ESOM) provides a faster convergence rate than NN-$K$ at the cost of one additional round of communication for the dual ascent step. Notably, replacing the augmented Lagrangian in the ESOM formulation with its linear approximation results in the EXTRA update equations, showing the relationship between both approaches.
    
    In some cases, computation of the Hessian is impossible because second order information is not available. Quasi-Newton methods like the Broyden-Flectcher-Goldman-Shanno (BFGS) algorithm approximate the Hessian when it cannot be directly computed. The distributed BFGS (D-BFGS) algorithm \cite{Eisen2017} replaces the second order information in the primal update in ESOM with a BFGS approximation (\textit{i.e.} replaces $D_i^{(k)}$ in a call to the Hessian approximation equations in Algorithm \ref{alg:NetNewtonPa} with an approximation), and results in essentially a ``doubly" approximate Hessian inverse. In \cite{Eisen2019} the D-BFGS method is extended so that the dual update also uses a distributed Quasi-Newton update scheme, rather than gradient ascent. The resulting primal-dual Quasi-Newton method requires two consecutive iterative rounds of communication doubling the communication overhead per primal variable update compared to its predecessors (NN-$K$, ESOM, and D-BFGS). However, the resulting algorithm is shown by the authors to still converge faster in terms of required communication.  
    

    \subsection{Convex Surrogate Methods}  
    While the approximate Newton methods in \cite{Mokhtari2016, Eisen2017, Eisen2019} optimize a quadratic approximation of the augmented Lagrangian of \eqref{eq:penalty_distributed_problem}, other distributed methods allow for more general and direct convex approximations of the distributed optimization problem. These convex approximations generally require the gradient of the joint objective function which is inaccessible to any single robot. In the NEXT family of algorithms \cite{di2016next} dynamic consensus is used to allow each robot to approximate the global gradient, and that gradient is then used to compute a convex approximation of the joint cost function locally. A variety of surrogate functions, $U(\cdot)$, are proposed including linear, quadratic, and block-convex which allows for greater flexibility in  tailoring the algorithm to individual applications. Using its surrogate of the joint cost function, each robot updates its local variables iteratively by solving its surrogate the problem, and then taking a weighted combination of the resulting solution with the solutions of its neighbors. To ensure convergence NEXT algorithms require a series of decreasing step-sizes, resulting in generally slower convergence rates as well as additional hyperparameter tuning.
    
   The SONATA \cite{sun2017distributed} algorithm extends the surrogate function principles of NEXT, and proposes a variety of non-doubly stochastic weighting schemes that can be used to perform gradient averaging similar to the push-sum protocols. The authors of SONATA also show that a several configurations of the algorithm result in already proposed distributed optimization algorithms including Aug-DGM, Push-DIG, and ADD-OPT.
    
 	\begin{algorithm}[t]
	\caption{NEXT}\label{alg:NEXT}
    \textbf{Private variables:} $\mc{P}_i = \left(x_i^{(k)}, \tilde{x}_i^{(k)}, \ti{\pi}_i^{(k)} \right)$
    \\\textbf{Public variables:} $\mc{Q}_i^{(k)} = \left(z_i^{(k)}, y_i^{(k)} \right)$
    \\\textbf{Parameters:} $\mc{R}_i^{(k)} = \left(\alpha^{(k)}, w_i, U(\cdot), \mc{K} \right)$
    \\\textbf{Update equations:}
    \begin{align*}
    x_i^{(k+1)} &= \sum_{j \in \mc{N}(i)} w_{ij} z_j^{(k)} \\
    y_i^{(k+1)} &= \sum_{j \in \mc{N}(i)} w_{ij} y_j^{(k)} + \left[ \nabla f_i(x_i^{(k + 1)}) -  \nabla f_i(x_i^{(k)}) \right] \\
    \ti{\pi}_i^{(k+1)} &= N \cdot y_i^{(k + 1)} - \nabla f_i(x_i^{(k+1)})\\
    \ti{x}_i^{(k+1)} &= \argmin_{x \in \mc{K}} U \left( x; x_i^{(k+1)}, \ti{\pi}_i^{(k+1)} \right)\\
    z_i^{(k+1)} &= x_i^{(k+1)} + \alpha^{(k)} \left(\ti{x}_i^{(k+1)} - x_i^{(k+1)} \right)
    \end{align*}
    \end{algorithm}   
    
    \subsection*{Applications of Distributed Sequential Convex Programming}
    Distributed sequential convex programming methods have been applied to pose graph optimization problems \cite{choudhary2017distributed} using a quadratic approximation of the objective function along with Gauss-Siedel updates to enable distributed local computations among the robots. The NEXT family of algorithms have been applied to a number of learning problems where data is distributed including semi-supervised support vector machines \cite{scardapane2016distributed}, neural network training \cite{scardapane2017framework}, and clustering \cite{altilio2019distributed}.
   
	\section{Alternating direction method of multipliers}\label{sec:ADMM}
	
    
    Considering the optimization problem in \eqref{eq:general_problem_with_agreement} with only agreement constraints, we have
    \begin{align}
        \min_{x_i \forall i \in \mathcal{V}} &\sum_{i \in \mathcal{V}} f_i(x_i)\\
        \text{subject to } & x_i = x_j  \qquad\forall(i, j) \in \mathcal{E}.
    \end{align}
    The \textit{method of multipliers} solves this problem by alternating between minimizing the augmented Lagrangian of the optimization problem with respect to the primal variables $x_1, \dots, x_n$ (the ``primal update") and taking a gradient step to maximize the augmented Lagrangian with respect to the dual (the ``dual update'').  In the \textit{alternating direction method of multipliers} (ADMM), given the separability of the global cost function, the primal update is executed as successive minimizations over each primal variable (\textit{i.e.} choose the minimizing $x_1$ with all other variables fixed, then choose the minimizing $x_2$, and so on). Most ADMM-based approaches do not satisfy our definition of distributed in that either the primal updates take place sequentially rather than in parallel or the dual update requires centralized computation \cite{terelius2011decentralized,houska2016augmented,chatzipanagiotis2015augmented, iutzeler2013asynchronous}.  However, the \textit{consensus alternating direction method of multipliers} (C-ADMM) provides an ADMM-based optimization method that is fully distributed: the nodes alternate between updating their primal and dual variable and communicating with neighboring nodes \cite{mateos2010distributed} .
    
    
    
    In order to achieve a distributed update of the primal and dual variables, C-ADMM alters the agreement constraints between agents with an existing communication link by introducing auxiliary primal variables to \eqref{eq:general_problem_with_agreement} (instead of the constraint $x_i = x_j$, we have two constraints: $x_i = z_{ij}$ and $x_j = z_{ij}$).  Considering the optimization steps across the entire network, C-ADMM proceeds by optimizing the auxiliary primal variables, then the original primal variables, and then the dual variables as in the original formulation of ADMM.  We can perform minimization with respect to the primal variables and gradient ascent with respect to the dual on an augmented Lagrangian that is fully distributed among the robots:
        \begin{align}
            \mathcal{L}_{a} &= \sum_{i \in \mc{V}}  f_{i}(x_{i}) +  y_{i}^{\top}x_i + \frac{\rho}{2}\sum_{j \in \mc{N}_i} \norm{x_{i} - z_{ij}}_{2}^{2}, \label{eq:lagrangian}
        \end{align}
    where $y_{i}$ represents the dual variable that enforces agreement between robot $i$ and its communication neighbors.  The parameter $\rho$ that weights the quadratic terms in $\mathcal{L}_a$ is also the step size in the gradient ascent of the dual variable. Furthermore, we can simplify the algorithm by noting that the auxiliary primal variable update can be performed implicitly ($z_{ij}^* = \frac{1}{2}\left(x_{i} + x_j\right)$).  
    
    Algorithm \ref{alg:C-ADMM} summarizes the update procedures for the local primal and dual variables of each agent. The update procedure for $x_{i}^{k+1}$ requires solving an optimization problem which might be computationally intensive for certain objective functions. To simplify the update complexity, the optimization can be solved inexactly using a linear approximation of the objective function such as DLM \cite{Ling2015} and \cite{chang2014multi, farina2019distributed} or a quadratic approximation using the Hessian such as DQM \cite{MokhtariDQM2016}.
    
    \begin{algorithm}[t]
	\caption{C-ADMM} \label{alg:C-ADMM}
    \textbf{Private variables:} $\mc{P}_i^{(k)} = y_i^{(k)}$
    \\\textbf{Public variables:} $\mc{Q}_i^{(k)} = x_i^{(k)}$
    \\\textbf{Parameters:} $\mc{R}_i^{(k)} = \rho$
    \\\textbf{Update equations:} \begin{align*} 
    y_i^{(k+1)} &= y_i^{(k)} + \rho\sum_{j \in \mc{N}_i} \left(x_i^{(k)} - x_j^{(k)}\right)\\
    x_i^{(k+1)} &= \argmin_{x_i} \Bigg\{f_i(x_i) + x_i^\top y_i^{(k+1)} \cdots \\ &\qquad + \frac{\rho}{2} \sum_{j \in \mc{N}_i} \left\Vert x_i - \frac{1}{2}\left(x_i^{(k)} + x_j^{(k)}\right)\right\Vert^2_{2} \Bigg\}
    \end{align*}
    \end{algorithm}
    
    C-ADMM as presented in Algorithm \ref{alg:C-ADMM} requires each robot to optimize over a local copy of the global decision variable $x$. However, many robotic problems have an fundamental structure that makes maintaining global knowledge at every individual robot unnecessary: each robot's data relate only to a subset of the global optimization variables, and each agent only requires a subset of the optimization variable for its role.  For instance, in distributed SLAM, a memory-efficient solution would require a robot to optimize only over its local map and communicate with other robots only messages of shared interest.  The SOVA method \cite{ola2020SOVA} leverages the separability of the optimization variable to achieve orders of magnitude improvement in convergence rates, computation, and communication complexity over C-ADMM methods. 
    
    In SOVA, each agent only optimizes over variables relevant to its data or role, enabling robotic applications in which agents have minimal access to computation and communication resources. SOVA introduces consistency constraints between each agent's local optimization variable and its neighbors, mapping the elements of the local optimization variables, given by
    \begin{align*}
        \Phi_{ij}x_{i} = \Phi_{ji}x_{j} \quad \forall j \in \mc{N}_{i},\ \forall i \in \mc{V}
    \end{align*}
    where $\Phi_{ij}$ and $\Phi_{ji}$ map elements of $x_{i}$ and $x_{j}$ to a common space. C-ADMM represents a special case of SOVA where $\Phi_{ij}$ is always the identity matrix. The update procedures for each agent reduce to
    the equations given in Algorithm \ref{alg:SOVA}.  While SOVA allows for improved 
	
	\begin{algorithm}[t]
	\caption{SOVA} \label{alg:SOVA}
    \textbf{Private variables:} $\mc{P}_i^{(k)} = y_i^{(k)}$
    \\\textbf{Public variables:} $\mc{Q}_i^{(k)} = x_i^{(k)}$
    \\\textbf{Parameters:} $\mc{R}_i^{(k)} = (\rho, \Phi)$
    \\\textbf{Update equations:} \begin{align*}
    y_i^{(k+1)} &= y_i^{(k)} + \rho\sum_{j \in \mc{N}_i} \Phi_{ij}^\top \left(\Phi_{ij}x_i^{(k)} - \Phi_{ji}x_j^{(k)}\right)
    \\ x_i^{(k+1)} &= \argmin_{x_i} \Bigg\{f_i(x_i) + x_i^\top y_i^{(k+1)} \cdots \\ &\qquad + \rho \sum_{j \in \mc{N}_i} \left\Vert \Phi_{ij} x_i - \frac{1}{2}\left(\Phi_{ij}x_i^{(k)} + \Phi_{ji}x_j^{(k)}\right)\right\Vert^2_{2} \Bigg\}
     \end{align*}
    \end{algorithm}
    
    \subsection*{Applications of C-ADMM}    
	ADMM has been applied to bundle adjustment and pose graph optimization problems which involve the recovery of the $3D$ positions and orientations of a map and camera \cite{zhang2017distributed, eriksson2016consensus, choudhary2015exactly}, informative path planning \cite{park2019distributed}. However, these works require a central node for the dual variable updates. Other works employ the consensus ADMM variant without a central node \cite{natesan2017distributed} with other notable applications in target tracking \cite{ola2020targettracking}, signal estimation \cite{mateos2010distributed}, task assignment \cite{ola2019task}, motion planning \cite{bento2013message}, online learning \cite{Xu2015}, and parameter estimation in global navigation satellite systems \cite{khodabandeh2019distributed}. Further applications of C-ADMM arise in trajectory tracking problems involving teams of robots using non-linear model predictive control \cite{ferranti2018coordination} and in cooperative localization \cite{kumar2016asynchronous}. Applications of SOVA include collaborative manipulation \cite{ola2020collab}. C-ADMM is adapted for online learning problems with streaming data in \cite{Xu2015}.

\section{Practical Notes}
\label{sec:notes_for_prac}
In Section \ref{sec:simulations}, we compare the performances of several distributed optimization methods on three benchmark problems that approximate the computational demands of typical robotics applications.  However, there are several significant considerations for translating an algorithm to a fast and efficient distributed solver.  Among the most important practical considerations (aside from choosing the most suitable algorithm) are parameter tuning, initialization, and communication graph topology.

\subsection{Communication Topology}

As a general rule, a more highly-connected graph facilitates faster convergence per communication iteration, regardless of the algorithm.  Just as the Fiedler value of a graph determines the rate of convergence in a pure consensus problem, the connectivity also has a direct effect on convergence in other distributed optimization problems.  Fully-connected communication enables fast optimization in a multi-robot network, and even makes centralized algorithms viable alternatives to distributed optimization, depending on computational constraints and the amount of local information possessed by each robot.  However, we are primarily concerned with the local, range-limited communication models that arise in practical large-scale robotics problems.

\subsection{Parameter Tuning}

The performance of each distributed algorithm that we consider is sensitive to the choice of parameters.  For instance, in DGD (Algorithm \ref{alg:DistGradDesc}), choosing $\alpha$ too large leads to divergence of the individual variables, while too small a value of $\alpha$ causes slow convergence. Similarly, C-ADMM (Algorithm \ref{alg:C-ADMM}) has a convergence rate that is highly sensitive to the choice of $\rho$, though convergence is guaranteed for all $\rho > 0$.  We study the sensitivity of the convergence rate to parameter choice in each simulation in Section \ref{sec:simulations}.  However, the optimal parameter choice for a particular example is not prescriptive for the tuning of other implementations.  Furthermore, while analytical results for optimal parameter selection are available for many of these algorithms, a practical parameter-tuning procedure is useful if an implementation does not exactly adhere to the assumptions in the literature.  Appendix \ref{sec:GSS} describes the Golden Section Search (GSS) algorithm, which provides a general (centralized) procedure for parameter tuning prior to deployment of a robotic system.

\subsection{Consensus Weights}

Several distributed optimization methods, including DGD, EXTRA, DDA, NN-$K$, and NEXT depend not only on stepsize parameters but also on consensus weights ($w$ in Algorithm \ref{alg:DistGradDesc}) by which each robot incorporates its neighbors' variables into its update equations.  In most cases, weights are assumed to be doubly stochastic (a robot's weights summed over its neighborhood is equal to one, as is the neighborhood's weights for the robot).  The specific choice of weights may vary depending on the assumptions made on what global knowledge is available to the robots on the network, as discussed thoroughly in \cite{xiao2004fast}.   For example, \cite{xiao2004fast} shows that finding pairwise symmetric weights that achieve the fastest possible consensus can be posed as a semidefinite program, which a computer with global knowledge of the network can solve efficiently.  However, we cannot always assume that global knowledge of the network is available, especially in the case of a time-varying topology. In most cases, Metropolis weights facilitate fast mixing without require global knowledge.  Each robot can generate its own weight vector after a single communication round with its neighbors.  In fact, Metropolis weights perform only slightly sub-optimally compared to centralized optimization-based methods \cite{jafarizadeh2010weight}:
	\begin{equation}
	w_{ij} = \begin{cases}
	        \frac{1}{\max\{\vert \mathcal{N}_i\vert, \vert \mathcal{N}_j \vert\} }& j \in \mathcal{N}_i \\ 1- \sum_{j^\prime \in \mathcal{N}_i} w_{ij^\prime} & i = j
	        \\ 0 & \text{else}
	\end{cases}
	\end{equation}
 Simulation results for algorithms with doubly stochastic mixing matrices use Metropolis weights.

	\section{Performance Comparisons in Case Studies}
	\label{sec:simulations}
	
	Many robotics problems have a distributed structure, although this structure might not be immediately apparent. 
	In many cases, applying distributed optimization methods requires reformulating the original problem into a separable form that allows for distributed computation of the problem variables locally by each robot. This reformulation often involves the introduction of additional problem variables local to each robot with an associated set of constraints relating the local variables between the robots. We provide three examples of distributed optimization to notable robotics problems in multi-drone vehicle target tracking, drone-robot coordinated package delivery, and multi-robot cooperative mapping.

	Our principal motivation in evaluating the distributed optimization methods described in Sections \ref{sec:DistGradDesc}-\ref{sec:ADMM} is to assess the suitability of each method in distributed robotics applications.  We measure the performances of a representative sample of the distributed optimization methods for three problem types: unconstrained least-squares, constrained convex optimization, and non-convex optimization. These classes of optimization problems are representative of a broad swath of robotic problems in estimation, localization, planning, and control.
	
	For a given distributed optimization method, important considerations for its include the generality of the algorithm (\textit{i.e.} the sensitivity of its performance across different scenarios to its parameters), the computational requirements (\textit{i.e.} each robot's computation time), and the communication requirements (\textit{i.e.} how many messages the robots must pass).  We use two important metrics in evaluating performance. The first is the normalized Mean Square Error (MSE) which is computed with respect to the optimal joint solution.
    When evaluating distributed optimization algorithms comparisons on the basis of MSE reduction per iteration can often skew results, and does not accurately reflect the total execution time of these algorithms. For example, methods like DIGing and NEXT require a multiple communication rounds, the execution time of which is not not accounted for in a per-iteration comparison. Similarly, methods like C-ADMM require local optimization steps at each iteration, which can significantly increase computation time depending on the cost function.  In robotics applications, both communication time and local computation time contribute to the overall execution time of these algorithms. Several works \cite{berahas2018balancing, reisizadeh2020fedpaq} propose application-specific cost metrics that assess both communication time and computation time.  In contrast to these works, we capture the trade-off between communication and computation with a single-parameter metric called Resource-Weighted Cost (RWC), which represents the total time required for a distributed optimization algorithm to converge below an acceptable MSE given some arbitrary communication and computation capability: 
    \begin{align}
        \label{eq:RWC}
        \text{RWC}_\lambda = \frac{t_{cp} + \lambda t_{cm}}{1 + \lambda} .
    \end{align}
	Here, $t_{cp}$ is the sum of the time spent by all robots performing local computations computed by summing processor time required for local update steps across the network. The $t_{cm}$ is the time spent by all robots communication with their neighbors which is approximated by counting the number of floating point values passed during the optimization. Finally, $\lambda$ is a scaling factor that weights communication to computation resource capability. As an example, a network of robots connected via a 5G cellular network would require significantly less time to communicate the same amount of data versus the same network of robots connected via a low power radio network. Sweeping across $\lambda$ corresponds to a pass over the range of possible network configurations and computational capabilities.

	\subsection{Distributed Multi-drone Vehicle Tracking}
	
	Linear least squares optimization, a convex unconstrained optimization problem, arises in a variety of robotics problems such as parameter estimation, localization, and trajectory planning. When extended to a multi-robot scenario, each of these problems can be reformulated as a distributed linear least squares optimization. In this simulation, we consider a distributed multi-drone vehicle target tracking problem in which robots connected by a communication graph, $\mc{G} = (\mc{V}, \mc{E})$, each record range-limited linear measurements of a moving target, and seek to collectively estimate the target's entire trajectory. We assume that each drone can communicate locally with nearby drones over the communication graph $\mc{G} = (\mc{V}, \mc{E})$.  The drones all share a model of the target's dynamics as \begin{equation}
	    x_{t+1} = A_t x_t + w_t,
	\end{equation} 
	where $x_t \in \mathbb{R}^4$ represents the position and velocity of the target in some global frame at time $t$, $A_t$ is a linear model of the targets dynamics, and $w_t \sim \mc{N}(0, Q_t)$ represents process noise (including the unknown control inputs to the target).  At every timestep when the target is sufficiently close to a drone $i$ (which we denote by $t \in \mc{T}_i$), that robot collects an observation according to the measurement model
	\begin{equation}
	    y_{i,t} = C_{i,t} x_t + v_{i, t} \: ,
	\end{equation} where $y_{i,t} \in \mathbb{R}^2$ is a positional measurement, $C_{i,t}$ is the measurement model of drone $i$, and $v_{i, t} \sim \mathcal{N}(0, R_{i, t})$ is measurement noise. All of the drones have the same model for the prior distribution of the initial state of the target $\mc{N}(\bar{x}_0, \bar{P}_0)$.
	The global cost function is of the form 
	\begin{equation}
	    \label{eq:trajectory_estimation}
	    \begin{aligned}
	    f(x) = & \norm{x_{0} - \bar{x}_{0}}_{\bar{P}_{0}^{-1}}^{2} + \sum_{t=1}^{T-1} \norm{x_{t+1} - A_t x_{t}}_{Q_t^{-1}}^{2}  \\
	    & + \sum_{i \in \mc{V}} \sum_{t \in \mc{T}_i} \norm{y_{i,t} - C_{i, t} x_{t}}_{R^{-1}}^{2},
	    \end{aligned}
	\end{equation}
	while the local cost function for drone $i$ is
	\begin{equation}
	    \label{eq:trajectory_estimation_local}
	    \begin{aligned}
	    f_i(x) = & \frac{1}{N}\norm{x_{0} - \bar{x}_{0}}_{\bar{P}_{0}^{-1}}^{2} + \sum_{t=1}^{T-1} \frac{1}{N} \norm{x_{t+1} - A_t x_{t}}_{Q_t^{-1}}^{2}  \\
	    & + \sum_{t \in \mc{T}_i} \norm{y_{i,t} - C_{i, t} x_{t}}_{R^{-1}}^{2}.
	    \end{aligned}
	\end{equation}
	
\begin{figure}[tb]
        \centering
        \includegraphics[width=\columnwidth]{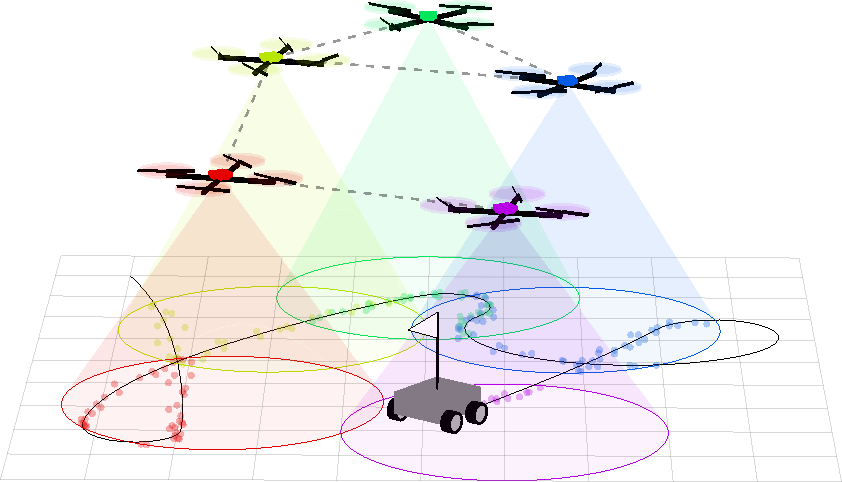}
        \caption{A visualization of the distributed multi-drone vehicle target tracking.  Each robot (colored quadrotor) records noisy observations when the target (flagged ground vehicle) is within its measurement range. The robots communicate over a network (dashed lines) to converge to the globally optimal trajectory estimate.}
        \label{fig:BLE_schematic}
    \end{figure}

	In our results, we consider only a batch solution to the problem (finding the full trajectory of the target given each robot's full set of measurements).  Methods for performing the estimate in real-time through filtering and smoothing steps have been well studied, both in the centralized and distributed case \cite{olfati2007distributed}. An extended version of this multi-robot tracking problem is solved with distributed optimization in \cite{ola2020targettracking}. A rendering of a representative instance of this multi-robot tracking problem is shown in Figure \ref{fig:BLE_schematic}.
		
\begin{figure}[tb]
	    \centering
	    \includegraphics[width=\linewidth]{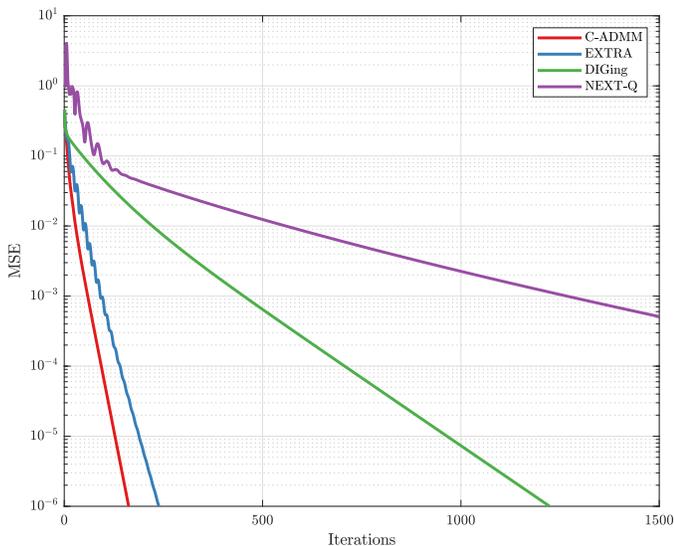}
	    \caption{MSE per iteration on a distributed multi-drone vehicle target tracking problem with $N=10$ and $n = 64$. }
	    \label{fig:BLE_cpi}
	\end{figure}

\begin{figure}[tb]
        \centering
        \includegraphics[width=\columnwidth]{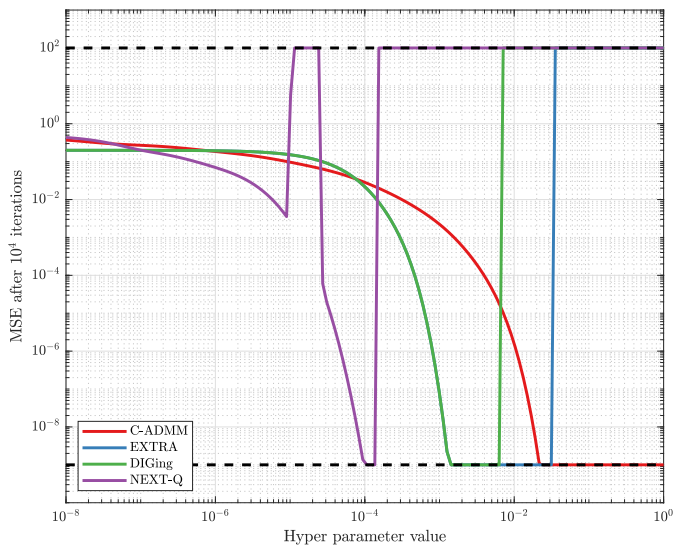}
        \caption{step-size hyper parameter sensitivity sweep on a distributed multi-drone vehicle target tracking problem for $N=10$ and $n=64$. The dashed lines are thresholds for divergence (top) and convergence (bottom) in terms of MSE after $10^4$ decision variable updates.}
        \label{fig:BLE_step_sense}
    \end{figure}

    In Figures \ref{fig:BLE_cpi} and \ref{fig:BLE_step_sense} several distributed optimization algorithms are compared on an instance of the distributed multi-drone vehicle tracking problem. For this problem instance, 10 simulated drones seek to estimate the target's trajectory over 16 time steps resulting in a decision variable dimension of $n = 64$. We compare four distributed optimization methods which we consider to be representative of the taxonomic classes outlined in the sections above: C-ADMM, EXTRA, DIGing, and NEXT-Q. Figure \ref{fig:BLE_cpi} shows that C-ADMM and EXTRA have similar fast convergence rates per iteration while DIGing and NEXT-Q are 4 and 15 times slower respectively to converge below an MSE of $10^{-6}$. The step-size hyperparameters for each method are computed by GSS (for NEXT-Q which uses a two parameter decreasing step-size we fix one according to the values recommended in \cite{di2016next}).
    
    As mentioned in Section \ref{sec:notes_for_prac}, tuning is essential for achieving robust and efficient convergence with most distributed optimization algorithms. Figure \ref{fig:BLE_step_sense} shows the sensitivity of these methods to variation in step-size, and highlights that three of the methods (all except ADMM) become divergent for certain subsets of the tested hyper parameter space.

\begin{figure}[tb]
        \centering
        \includegraphics[width=\columnwidth]{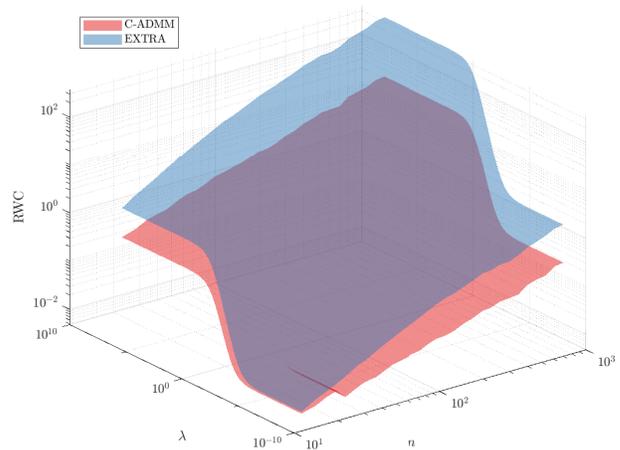}
        \caption{Resource-weighted cost comparison of C-ADMM and EXTRA on the distributed multi-drone vehicle tracking problem. GSS was used to find the best step-sizes for each of the algorithms for each problem instance. Computation time was measured in MATLAB on a single CPU core.}
        \label{fig:BLE_surf}
    \end{figure}

    While MSE reduction per iteration and hyper parameter sensitivity are useful for understanding the performance of these methods on this specific problem instance, they do not provide information about the scalability of these methods. Specifically, how do these algorithms scale on real networks where both communication and computation overhead are important? To answer this question we look to the RWC (\ref{eq:RWC}) which provides a more comprehensive analysis of the problem. In Figure \ref{fig:BLE_surf} we show how the two fastest converging algorithms from the single instance analysis, C-ADMM and EXTRA, perform with a sweep across both problem size and RWC parameter $\lambda$. 
    
    The RWC surfaces in Figure \ref{fig:BLE_surf} show that C-ADMM has a lower RWC compared to EXTRA at all points in the sweep, except where both $\lambda$ and $n$ are small where the algorithms have roughly the same RWC. This corresponds to scenarios with relatively small decision variables, and a network configuration where computation is costly and communication is cheap.
    
	\subsection{Multi Drone-Robot Coordinated Package Delivery}
	Many robotics scenarios, especially in planning and control, involve constrained optimization problems, in which the decision variable must satisfy constraints such as control limits, state bounds, or load limits. In general, these problems are of the form
	\begin{equation}
	    \label{eq:constrained_convex}
	    \begin{aligned}
	    &\underset{x}{\text{minimize}} &&\sum_{i=1}^{N} f_{i}(x) \\
	    &\text{subject to} &&g(x) \leq 0 \\
	                    &  &&h(x) = 0.
        \end{aligned}
	\end{equation}
	We consider the case for which the constraints $g(x) \le 0$ and $h(x) = 0$ are convex, and each local cost function $f_{i}(x)$ is convex. Note that if $g(x)$ is a convex function, then $g(x)\le 0$ is a convex constraint, and $h(x) = 0$ is a convex constraint if and only if $h(x)$ is an affine function.  

    \begin{figure}[t]
        \centering
        \includegraphics[width=0.95\columnwidth]{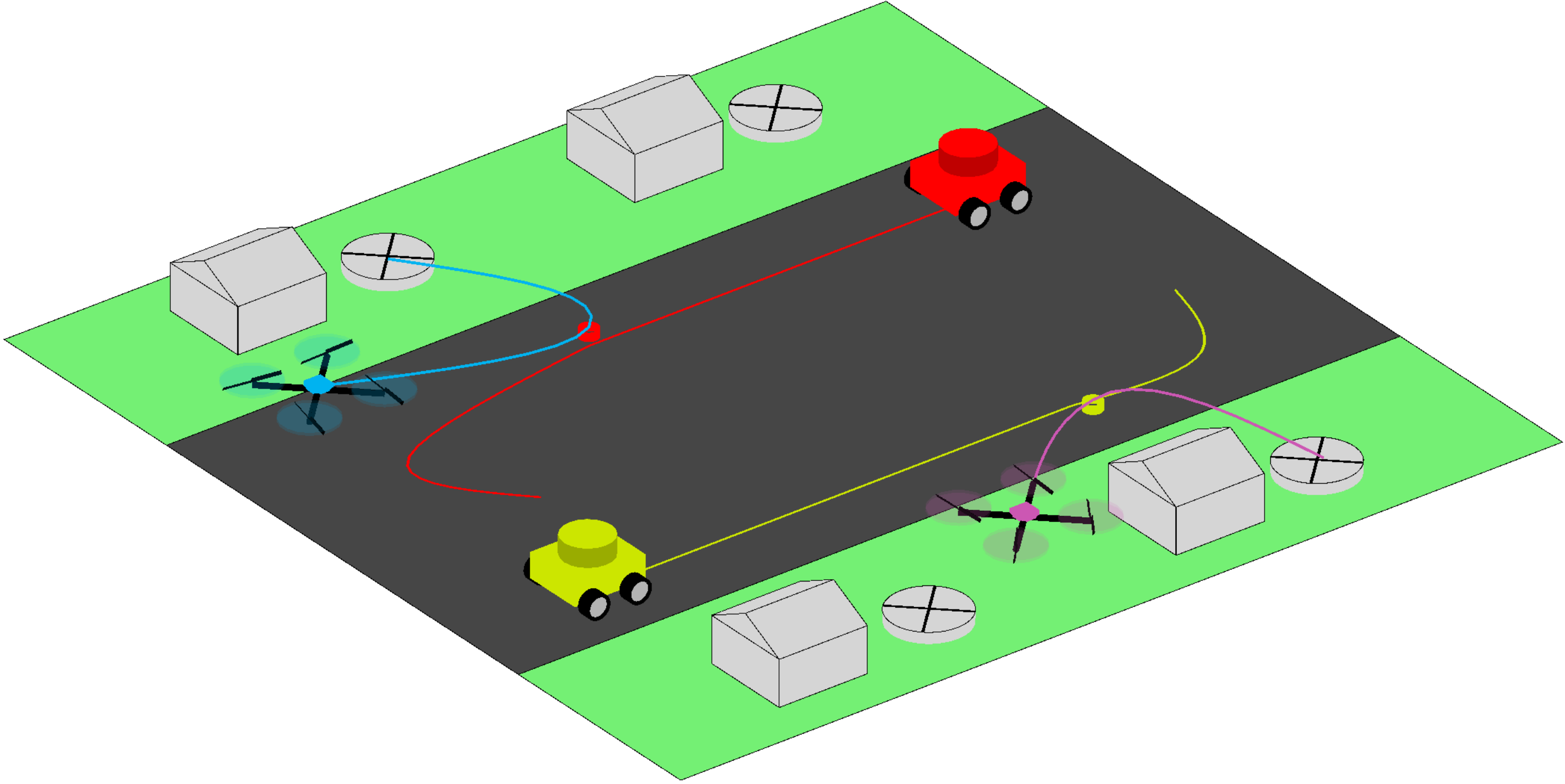}
        \caption{Coordinated package delivery by aerial and ground robots. The aerial robots pick up packages from the ground robots for delivery to areas inaccessible to the ground robots which remain within the inner rectangular region.}
        \label{fig:robot_delivery}
    \end{figure}
	
	
	The distributed subgradient methods (\cite{nedic2009}, \cite{shi2015extra}, \cite{nedic2014distributed}, \cite{nedic2017digging} \cite{jakovetic2014fast}) do not handle constrained optimization problems, and extending these methods is beyond the scope of this paper. Consequently, we compare C-ADMM to dual averaging with push-sum (PS-DA) \cite{tsianos2011distributed} and the distributed convex approximation method with consensus on the network gradients (NEXT-Q) in which we take a quadratic approximation of the objective function using the problem Hessian.
	
	Consider $N$ aerial robots delivering packages within a remote area. Each robot meets with a limited number of ground robots $M$ at a given time to pick up packages for delivery to specified locations. In addition, the ground robots are constrained to move within certain zones. Depending on the size of the ground robots, these constraints are included to keep the robots on the roads (for larger robots) or on sidewalks (for smaller robots). We want to compute a trajectory for each robot that minimizes its energy consumption and satisfies the package pick-up, control, and state constraints. The resulting optimization problem is
	\begin{equation}
	    \label{eq:package_delivery}
	    \begin{aligned}
	    &\underset{x_{a},u_{a},x_{g},u_{g}}{\text{minimize}} &&\sum_{i=1}^{N} u_{a,i}^{T}Q_{a,i}u_{a,i} + \sum_{j=1}^{M} u_{g,j}^{T}Q_{g,j}u_{g,j} \\
	    &\text{subject to} &&x_{a,i} = x_{g,j} \quad \forall t \in \mc{T}_{i,j} \ \forall i, \ \forall j \\
	    &                  &&f_{a}(x_{a,i},u_{a,i}) = 0 \quad \forall i \\
	    &                  &&f_{g}(x_{g,j},u_{g,j}) = 0 \quad \forall j \\
	    &                  &&h(x_{g,j},u_{g,j}) \leq 0 \quad \forall j
	    \end{aligned}
	\end{equation}
	where $x_{a}$ and $u_{a}$ denote the state and control inputs of the aerial robots while $x_{g}$ and $u_{g}$ denote the state and control inputs of the ground robots over the duration of the package delivery problem. $Q_{a}$ and $Q_{g}$ are positive definite weight matrices on the control inputs of the aerial and ground robots. The constraint in \eqref{eq:package_delivery} ensures the aerial robots meet with the ground robots at the specified times in $\mc{T}_{i,j}$. The robots' states follow the dynamics represented by $f(\cdot)$. Closed-form solutions for the constrained convex optimization problem \eqref{eq:package_delivery} do not exist. 

	With ${x_{a} \in \mathbb{R}^{6}}$ and ${x_{g} \in \mathbb{R}^{4}}$, we impose convex state constraints on the position and velocities of the ground robots, representing constraints on the zones which the robots can occupy. In addition, we constrain the control inputs of the ground robots and assume affine dynamics for the ground and aerial robots. The robots' delivery assignments begin and end at specified stations which we include as constraints in the optimization problem. 
	
	Representing the trajectory of all the robots as $Z$ including their states and control inputs, we define the mean square error (MSE) of each robot's solution to the joint solution as
	\begin{equation}
	    \text{MSE}(Z_{1},\cdots,Z_{N}) = \frac{1}{N} \sum_{i = 1}^{N} \norm{Z_{i} - Z^{\star}}_{2}^{2}
	\end{equation}
	where $Z^{\star}$ represents the joint solution of the problem.
	
\begin{figure}[t]
        \centering
        \includegraphics[width=0.95\columnwidth]{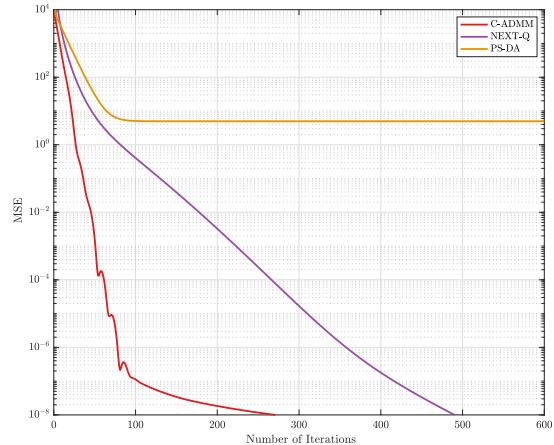}
        \caption{Convergence error of C-ADMM, NEXT-Q, and PS-DA for the package delivery problem on range-limited graphs. PS-DA converges slowly compared to C-ADMM and NEXT-Q.}
        \label{fig:coordinated_delivery_convergence}
    \end{figure}

	We examine the performance of C-ADMM, PS-DA, and NEXT-Q on the coordinated package delivery problem using the MSE. We show the joint solution in Figure \ref{fig:robot_delivery} where the aerial robots pick up packages from the ground robots at locations indicated by the colored packages for delivery next to the homes inaccessible to the ground robots. The aerial robots conclude each delivery assignment at the marked locations indicated by the cross-hairs next to the homes. Similar to gradient descent methods, PS-DA shows notable sensitivity to the step-size and becomes unbounded for some values of the step-size. We selected the step-size that provided the fastest convergence. From Figure \ref{fig:coordinated_delivery_convergence}, C-ADMM converges faster to the joint solution compared to PS-DA and NEXT-Q which converges quickly once the agents obtain a good estimate of the problem gradients through consensus. PS-DA converges more slowly compared to the other methods.
	
\begin{figure}[t]
        \centering
        \includegraphics[width=0.95\columnwidth]{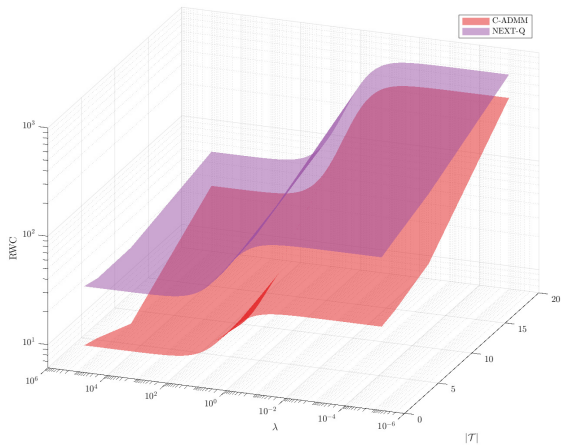}
        \caption{Resource-weighted cost for the package delivery problem on range-limited graphs. C-ADMM attains a lower total cost considering the computation and communication required by the robots.}
        \label{fig:coordinated_delivery_cost}
    \end{figure}

	In Figure \ref{fig:coordinated_delivery_cost}, we show the relative trade-off between the computation and communication overhead required by C-ADMM and NEXT-Q as the number of meeting constraints between the aerial and ground robots increases. We omit the cost of PS-DA from this figure as PS-DA converges significantly more slowly compared to the other methods. The constrained problem becomes increasingly difficult for an increasing number of meeting constraints, as reflected in Figure \ref{fig:coordinated_delivery_cost}. Likewise, solving the constrained problem requires significant computation effort, and thus reducing the weight on the contribution of the computation effort to the resource-weight cost results in a lower resource-weighted cost. From Figure \ref{fig:coordinated_delivery_cost}, C-ADMM achieves a lower resource-weighted cost compared to NEXT-Q.
	
	\subsection{Cooperative Multi-robot Mapping}
	
\begin{figure}[t]
        \centering
        \includegraphics[width=\columnwidth]{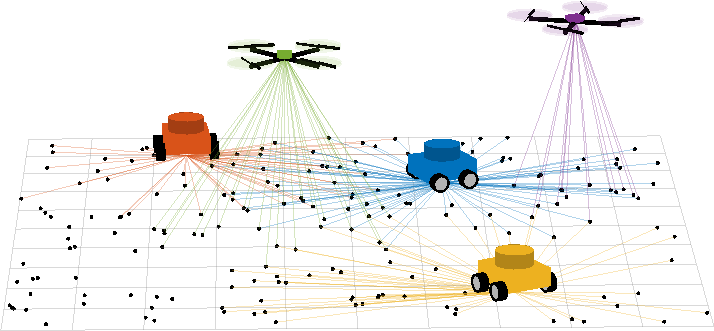}
        \caption{A visualization of the range-only distributed cooperative multi-robot mapping problem. Heterogeneous robots record noisy range measurements to landmarks and cooperatively estimate the landmarks' positions.}
        \label{fig:Scenario3Visualization}
    \end{figure}

	In our third evaluation of distributed methods for robotics, we consider distributed nonlinear, nonconvex optimization. Many problems in robotics in robotics require optimization over nonconvex.  For example, several recent papers have addressed distributed approaches to SLAM with problem-specific algorithms.  In this work, we consider a milder (though still nonconvex) problem: distributed cooperative multi-robot mapping of labeled landmarks using range-only measurements.  The distributed cooperative multi-robot mapping, visualized in Figure \ref{fig:Scenario3Visualization}, consists of $n$ robots navigating around an environment while taking measurements of the distance between their known positions and the unknown positions of $m$ landmarks.  The separable cost function with agreement constraints is given as:
	
	\begin{equation}
	\begin{aligned}
	    \min_{x_{11}, \dots, x_{nm}} &\sum_{i = 1}^n  \sum_{k = 1}^m \sum_{t \in \mc{T}_{ik}} \frac{w_{ik}^2}{2} \left(\norm{p_i - x_{ik}} - d_{ik}(t)\right)^2
	    \\\text{subject to } &x_{ik} = x_{jk} \quad \forall k, \forall j \in \mathcal{N}_i,
	\end{aligned}
	\label{eq:distmapping}
	\end{equation}
	where $x_{ik}$ denotes the $i$th robot's estimate of the location of the $k$th landmark.

	
	


    Although the distributed methods that we consider do not have convergence guarantees for non-convex optimization problems, we have evaluated each algorithm based on the most immediate interpretation of the algorithms.  In particular, we consider EXTRA and C-ADMM for this problem.  The application of DGD methods to differentiable but non-convex problems such as \eqref{eq:distmapping} is immediate, although there is no guarantee of convergence to the global minimum.  
    
    An important consequence of solving a nonlinear least squares problem to C-ADMM is that the primal update step of the algorithm no longer consists of a single linear update.  In this implementation, each robot solves the minimization step to completion 
	(up to a specified error tolerance) before communicating with its neighbors, incurring additional computational cost per iteration.  As a result, the C-ADMM update equations between communication iterations can require significantly more computation than the EXTRA update equations.  Despite the higher computation cost per iteration, empirical results suggest that C-ADMM still maintains a distinct advantage over EXTRA.
	
	
	
	\begin{figure}[t]
        \centering
        \includegraphics[width=\columnwidth]{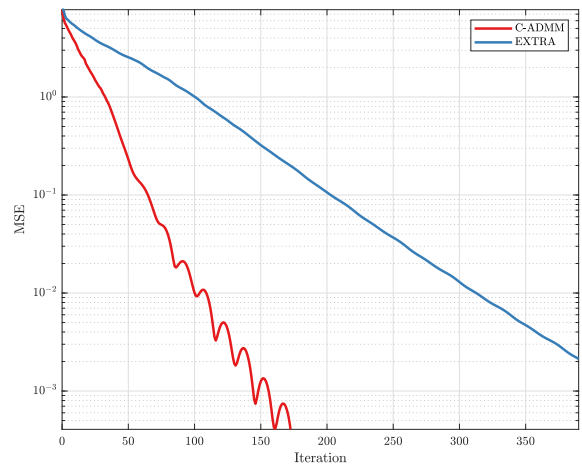}
        \caption{Mean-square-error of robots' estimates as a function of iteration for C-ADMM and EXTRA for a fixed graph containing 20 nodes and optimal choices of step-size parameters.}
    \end{figure}
	
	\begin{figure}[t]
        \centering
        \includegraphics[width=\columnwidth]{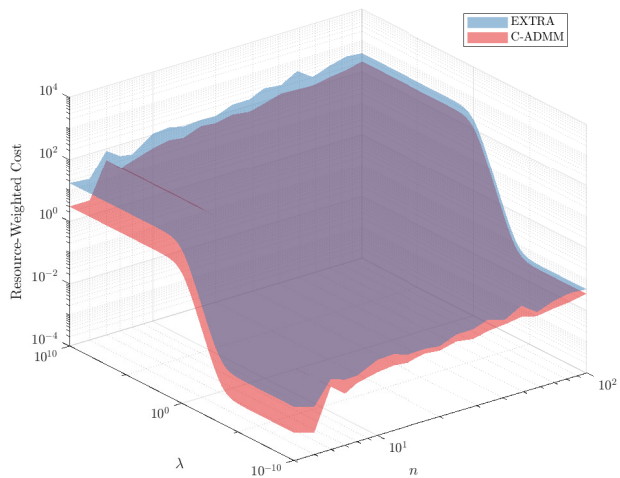}
        \caption{Resource weighted cost of distributed optimization algorithms on a distributed cooperative multi-robot mapping problem. }
    \end{figure}

\section{Open Problems in Distributed Optimization for Robotics}
\label{sec:open_problems}
Distributed optimization methods have primarily focused on solving unconstrained convex optimization problems, which constitute a notably limited subset of robotics problems. Generally, robotics problems involve non-convex objectives and constraints, which render these problems not directly amenable to many existing distributed optimization methods. For example, problems in multi-robot motion planning, SLAM, learning, distributed manipulation, and target tracking are often non-convex and/or constrained. 
    
Both DGC methods and C-ADMM methods can be modified for non-convex and constrained problems, however few examples of practical algorithms or rigorous analyses of performance for such modified algorithms exist in the literature.  Specifically, while C-ADMM is naturally amenable to constrained optimization, there are many possible ways to adapt C-ADMM to non-convex objectives, which have yet to be explored.  One way to implement C-ADMM for non-convex problems is for each primal step to solve itself a non-convex optimization (e.g. through a quasi-Newton method, or interior point method).  Another option is to perform successive quadratic approximations in an outer loop, and use C-ADMM to solve each resulting quadratic problem in an inner loop.  The trade-off between these two options has not yet been explored in the literature, especially in the context of non-convex problems in robotics.

Likewise, many distributed optimization methods do not consider communication between agents as an expensive resource, given that many of these methods were developed for problems with reliable communication infrastructure (e.g. multi-core computing, or computing in a hard-wired cluster). However, communication takes on greater prominence in robotics problems as robots often operate in regions with limited communication infrastructure. The absence of reliable communication infrastructure also leads to communication delays and dropped message packets.  This highlights the need for research analyzing the robustness of distributed optimization methods to unreliable communication networks, and the development of new algorithms robust to such real-world communication faults. 

Another valuable direction for future research is in developing algorithms specifically for computationally limited robotic platforms, in which the timeliness of the solution is as important as the solution quality.  In general, many distributed optimization methods involve computationally challenging procedures that require significant computational power, especially distributed methods for constrained problems. These methods ignore the significance of computation time, assuming that agents have access to significant computational power. These assumptions often do not hold in robotics problems. Typically, robotics problems unfold over successive time periods with an associated optimization phase at each step of the problem. As such, agents must compute their solutions fast enough to proceed with computing a reasonable solution for the next problem which requires efficient distributed optimization methods.  Developing such algorithms specifically for multi-robot systems is an interesting topic for future work.

Finally, there are very few examples of distributed optimization algorithms implemented and running on multi-robot hardware.  This leaves a critical gap in the existing literature, as the ability of these algorithms to run efficiently and robustly on robots has still not be thoroughly proven.  In general, the opportunities for research in distributed optimization for multi-robot systems are plentiful.  Distributed optimization provides an appealing unifying framework from which to synthesize solutions for a large variety of problems in multi-robot systems.
    

\section{Conclusion}
\label{sec:conclusion}
The field of distributed optimization provides a variety of algorithms that can address important problems for multi-robot systems.  We have categorized distributed optimization methods into three broad classes---distributed gradient descent, distributed sequential convex programming, and the alternating direction method of multipliers (ADMM). We have presented a general framework for applying these methods to multi-robot scenarios, with examples in distributed multi-drone target tracking, multi-robot coordinated package delivery, and multi-robot cooperative mapping.  Our empirical simulation results suggest that C-ADMM provides an especially attractive algorithm for distributed optimization in robotics problems. While distributed optimization techniques can immediately apply to several fundamental applications in robotics, important challenges remain in developing distributed algorithms for constrained, non-convex robotics problems, and algorithms tailored to the limited computation and communication resources of robot platforms.

	\bibliographystyle{IEEEtran}
	\bibliography{references}

    \section{Appendix}
    \label{sec:appendix}
    
    \subsection{Fixed Step-size DGD General Form} \label{sec:FixedStepSize}
    As demonstrated in \cite{sundararajan2019canonical}, a range of fixed-step-size DGD methods, including EXTRA \cite{shi2015extra}, NIDS \cite{li2019decentralized}, Exact Diffusion \cite{yuan2018exact}, and DIGing \cite{nedic2017achieving} can be alternatively represented in a canonical form.  In this canonical form, weights define the inclusion of several terms common to fixed-step-size DGD including the local gradient and the local decision variable history. Algorithm \ref{alg:FixedStepSizeDGD} provides a unifying template for fixed step-size distributed gradient descent (DGD) methods with the selection of the parameters $\alpha,\ \zeta_0,\ \zeta_1,\ \zeta_2$, and $\zeta_3$ defining a specific DGD method. For an illustrative example, we obtain EXTRA (refer to Algorithm \ref{alg:EXTRA}) from the parameters ${\zeta_{0} = \frac{1}{2}}$, ${\zeta_{1} = 1}$, ${\zeta_{2} = 0}$, and ${\zeta_{3} = 0}$.
    	
	\begin{algorithm}[t]
	\caption{Fixed Step-Size Distributed Gradient Descent}\label{alg:FixedStepSizeDGD}
    \textbf{Private variables:} $\mc{P}_i = \emptyset$
    \\\textbf{Public variables:} $\mc{Q}_i^{(k)} = \left(x_i^{(k)}, z_i^{(k)}\right)$
    \\\textbf{Parameters:} $\mc{R}_i^{(k)} = (\alpha, \zeta_0, \zeta_1, \zeta_2, \zeta_3)$
    \\\textbf{Update equations:} \begin{align*}
    x_i^{(k+1)} &= x_i^{(k)} + \zeta_0 z_i^{(k)} - \zeta_1 \left(x_{i}^{(k)} - \sum_{j \in \mc{N}_i \cup \{i\}} w_{ij} x_j^{(k)}\right) \cdots \\
    &\qquad + \zeta_2 \left(z_{i}^{(k)} - \sum_{j \in \mc{N}_i \cup \{i\}} w_{ij} z_{j}^{(k)}\right) \cdots \\
    &\qquad - \alpha \nabla f_i\left((1 - \zeta_{3})x_i^{(k)} + \zeta_3 \sum_{j \in \mc{N}_i \cup \{i\}} w_{ij} x_j^{(k)} \right)\\
    z_i^{(k+1)} &= z_{i}^{(k)} - x_{i}^{(k)} + \sum_{j \in \mc{N}_i\cup\{i\}} w_{ij} x_j^{(k)}
    \end{align*}
    \end{algorithm}
    
    \subsection{Golden Section Search}\label{sec:GSS}
    We assume that rather than tuning parameters online using a distributed algorithm, the roboticist selects suitable parameters for an implementation before deploying a system, either using analytical results or simulation.  In this case, the most general (centralized) procedure for parameter tuning involves comparing the convergence performance of the system on a known problem for different parameter values.  However, while a uniform sweep of the parameter space may be effective for small problems or parameter-insensitive methods, it is not computationally efficient.  Given the convergence rate of a distributed method at particular choices of parameter, \textit{bracketing} methods provide parameter selections to more efficiently find the convergence-rate-minimizing parameter. For instance, Golden Section Search (GSS) provides a versatile approach for tuning a scalar parameter \cite{press1989numerical}.  Assuming that the dependence of convergence rate on the given parameter is strictly unimodal (as demonstrated in Section \ref{sec:simulations}), GSS maintains four parameter candidates and refines its list of candidates based on performance of the interior candidates.  Algorithm \ref{alg:GSS} outlines the basic procedure for parameter search.  We present the performance results in Section \ref{sec:simulations} according to the best parameter choice using GSS on $\log_{10}(a)$ for each parameter $a$ (typically requiring a total of $10$-$15$ problem evaluations).
	\begin{algorithm}[t]
	\caption{Golden Section Search} \label{alg:GSS}
    \begin{algorithmic}[1]
	\Function{GSS}{$a_0$, $a_3$, $f(\cdot)$}
	\State $h \gets a_3 - a_0$
	\State $(a_1, a2) \gets (a_0 + \varphi^2 h, a_0 + \varphi h)$
	\While{stopping criterion is not satisfied}
	    \If{$f(a_1) < f(a_2)$}
            \State $(a_3, a_2) \gets (a_2, a_1)$
            \State $h \gets \varphi h$
            \State $a_1 \gets a_0 + \varphi^2 h$
        \Else
            \State $(a_0, a_1) \gets (a_1, a_2)$
            \State $h \gets \varphi h$
            \State $a_2 \gets a_0 + \varphi h$
        \EndIf
	\EndWhile
	\EndFunction
	\end{algorithmic}
    \end{algorithm}

\end{document}